\documentclass[10pt,twocolumn,letterpaper]{article}

\usepackage{iccv}
\usepackage{times}
\usepackage{epsfig}
\usepackage{graphicx}
\usepackage{amsmath}
\usepackage{amssymb}

\usepackage{adjustbox}
\usepackage{multirow}
\usepackage{subcaption}
\usepackage{multirow,bigdelim}
\usepackage{booktabs}

\usepackage[pagebackref=true,breaklinks=true,letterpaper=true,colorlinks,bookmarks=false]{hyperref}

\iccvfinalcopy 

\def\iccvPaperID{350} 

\ificcvfinal\fi
\begin{document}

\title{Multi-modality Latent Interaction Network for Visual Question Answering}

\author{Peng Gao\textsuperscript{1}, \ Haoxuan You\textsuperscript{3}, \ Zhanpeng Zhang\textsuperscript{2},\\ Xiaogang Wang\textsuperscript{1},\ Hongsheng Li \textsuperscript{1} \\ \textsuperscript{1}CUHK-SenseTime Joint Lab, The Chinese University of Hong Kong\\
\textsuperscript{2}SenseTime Research \
\textsuperscript{3}Tsinghua University\\
\textit {\{1155102382@link, xgwang@ee, hsli@ee\}.cuhk.edu.hk}\\
}

\maketitle

\begin{abstract}
   Exploiting relationships between visual regions and question words have achieved great success in learning multi-modality features for Visual Question Answering (VQA). However, we argue that existing methods \cite{lu2016hierarchical} mostly model relations between individual visual regions and words, which are not enough to correctly answer the question. From humans' perspective, answering a visual question requires understanding the summarizations of visual and language information.
   In this paper, we proposed the Multi-modality Latent Interaction module (MLI) to tackle this problem. The proposed module learns the cross-modality relationships between latent visual and language summarizations, which summarize visual regions and question into a small number of latent representations to avoid modeling uninformative individual region-word relations. The cross-modality information between the latent summarizations are propagated to fuse valuable information from both modalities and are used to update the visual and word features. Such MLI modules can be stacked for several stages to model complex and latent relations between the two modalities and achieves highly competitive performance on public VQA benchmarks, VQA v2.0 \cite{balanced_vqa_v2} and TDIUC \cite{kafle2017analysis}. In addition, we show that the performance of our methods could be significantly improved by combining with pre-trained language model BERT\cite{devlin2018bert}.

\end{abstract}

\section{Introduction}

\begin{figure}[t]
        \begin{center}
                \includegraphics[width=\linewidth]{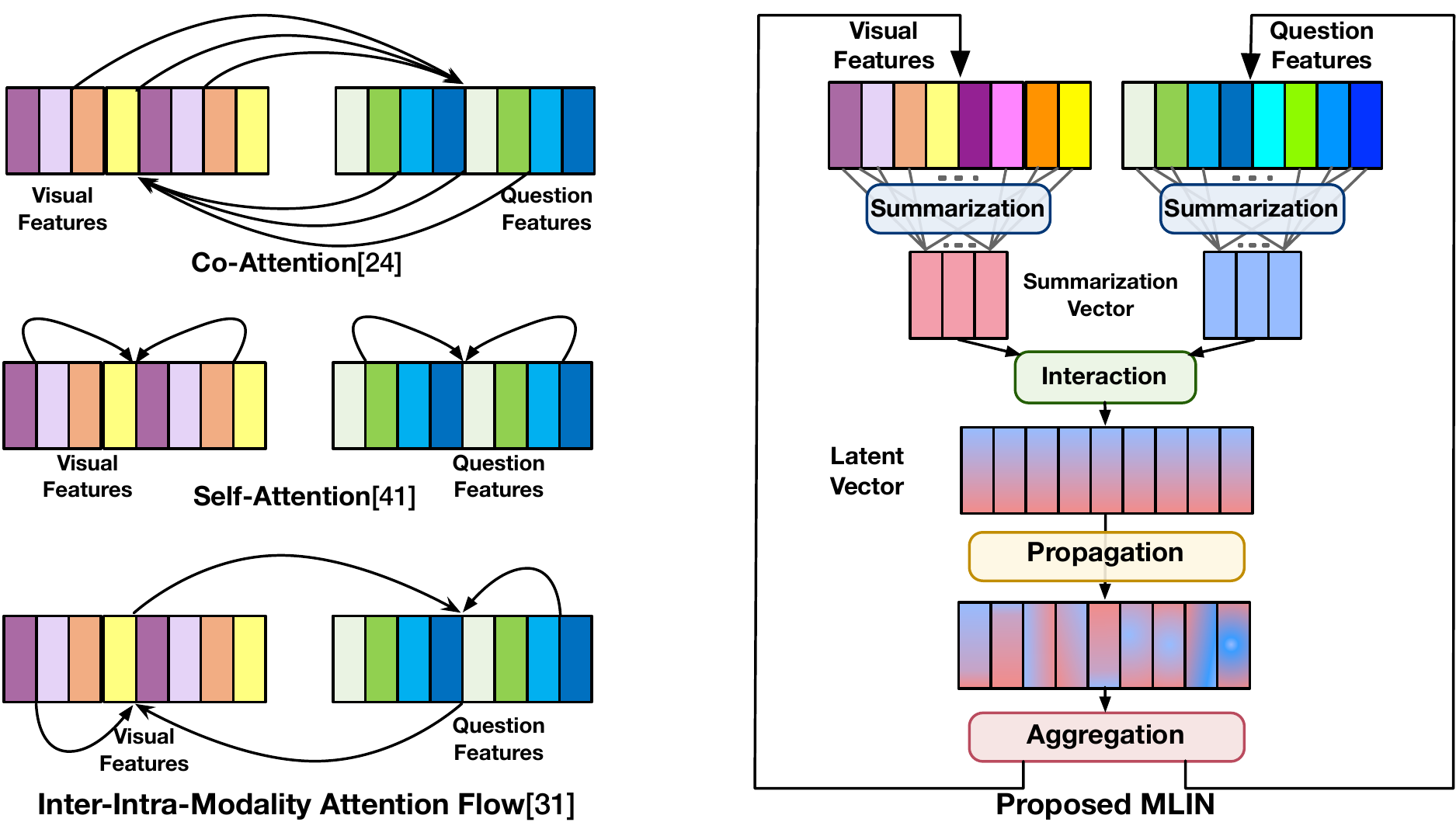}
        \end{center}
        \caption{Illustration of the information flow in our proposed MLI compared with previous approaches, namely, co-attention~\cite{lu2016hierarchical}, self-attention~\cite{vaswani2017attention} and intra-inter modality attention(DFAF)~\cite{peng2018dynamic}. Left side of each image represent visual feature while right side stands for question features.}
        \label{fig:comparison}
\end{figure}

Visual Question Answering~\cite{antol2015vqa, balanced_binary_vqa, balanced_vqa_v2} has received increasing attention from the research community. Previous approaches solve the Visual Question Answering (VQA) by designing better features~\cite{krizhevsky2012imagenet, simonyan2014very, he2016deep, huang2017densely, anderson2018bottom}, better bilinear fusion approaches~\cite{gao2016compact, fukui2016multimodal, kim2016hadamard, ben2017mutan, yu2018beyond} or better attention mechanisms~\cite{xu2015show, lu2016hierarchical, yang2016stacked, vaswani2017attention, peng2018dynamic}. Recently, relational reasoning has been explored for solving VQA and significantly improved performance and interpretability of VQA systems.

Despite relationships has been extensively adopted in different tasks, such as object detection~\cite{hu2018relation}, language modelling~\cite{devlin2018bert}, image captioning~\cite{yao2018exploring} and VQA~\cite{peng2018dynamic, geng20192nd}. Relational approaches for VQA were only proposed for modelling relationship between words and visual regions. Thus, relational reasoning requires large GPU memories because it needs to model relations between every pair. For VQA, modeling relationships between individual words and visual regions is not enough to correctly answer the question.

To model more complex cross-modality relations, we propose a novel Multi-modality Latent Interaction Network (MLIN) with MLI modules. Different from existing relational VQA methods, the MLI module first encodes question and image features into a small number of latent visual and question summarizaiton vectors. Each summarization vector can be formulated as the weighted pooling over visual or word features, which summarizes certain aspect of each modality from a global perspective and therefore encodes richer information compared with individual word and region features. After acquiring summarizations for each modality, we establish visual-language associations between the multi-modal summarization vectors and propose to propagate information between summarization vectors to model the complex relations between language and vision. Each original visual region and word feature would finally aggregate information from the updated latent summarizations using attention mechanisms and residual connections to predict the correct answers. 

Our proposed MLIN achieves competitive performance on VQA benchmarks, including VQA v2.0~\cite{balanced_vqa_v2} and TDIUC~\cite{kafle2017analysis}. In addition, we experiment how to combine pre-trained language model BERT~\cite{devlin2018bert} to improve VQA models. After integrating with BERT~\cite{devlin2018bert}, MLIN achieves better performance compared with state-of-the-art models.

Our proposed MLIN is related to the attention-based approaches. An illustration between previous approaches can be seen from Figure~\ref{fig:comparison}. Previous attention approaches that aggregate information can be classified into the following categories: (1) The co-attention mechanism~\cite{lu2016hierarchical} aggregates information from the other modality. (2) Transformer~\cite{vaswani2017attention} aggregates information inside each modality using key-query attention mechanism. (3) The intra- \& inter-modal attention(DFAF)~\cite{peng2018dynamic} propagate and aggregate information within and across multiple modalities. For intra-modality feature aggregation, attention is dynamically modulated by the other modality using the pooled features. Compared with previous approaches, MLIN does not aggregate features just from the large number of individual visual-word pairs but from the small number of multi-modal latent summarization vectors, which can capture high-level visual-language interactions with much smaller modal capacity.

Our contributions can be summarized into two-fold. 
(1) We propose the MLIN for modelling multi-modality interactions via a small number of multi-modal summarizations, which helps encode the relationships across modalities from global perspectives and avoids capturing too much uninformative region-word relations. (2) We carried out extensive ablation studies over each components of MLIN and achieve competitive performance on VQA v2.0~\cite{balanced_vqa_v2} and TDIUC~\cite{kafle2017analysis} benchmarks. Besides, we provide visualisation of our LMIN and have a better understanding about the interactions between multi-modal summarizations. We also explore how to effectively integrate the pre-trained language model ~\cite{devlin2018bert} into the proposed framework for further improving the VQA accuracy.

\section{Related Work}

\subsection{Representation Learning}
Learning good representations have been the foundations for advancing vision and Natural Language Processing (NLP) research. For computer vision, AlexNet~\cite{krizhevsky2012imagenet}, VGGNet~\cite{simonyan2014very}, ResNet~\cite{he2016deep} and DenseNet~\cite{huang2017densely} features achieved great success on image recognition~\cite{deng2009imagenet}. For NLP, word2vec~\cite{mikolov2013distributed}, GloVe~\cite{pennington2014glove}, Skipthough~\cite{kiros2015skip}, ELMo~\cite{peters2018deep}, GPT~\cite{radford2019language}, VilBERT~\cite{lu2019vilbert} and BERT~\cite{devlin2018bert} achieved great success at language modelling. The successful representation learning in vision and language has much benefitted multi-modality feature learning. Furthermore, bottom-up \& top-down features~\cite{anderson2018bottom} for VQA and image captioning greatly boosted the performance of multi-modality learning based on the additional visual region  (object detection~\cite{ren2015faster}) information.
\subsection{Relational Reasoning}
Our work is mostly related to the relational reasoning approaches. Relational reasoning approaches try to solve VQA by learning the relationships between individual visual regions and words. Co-attention based~\cite{lu2016hierarchical} approaches can be seen as modelling the relationship between each word and visual region pairs using the attention mechanism. Transformer~\cite{vaswani2017attention} proposed to use the key-query-value attention mechanism to model the relationship inside each modality. Simple relational networks~\cite{santoro2017simple, hu2019weakly} reason over all region pairs in the image by concatenating region features. Besides VQA, relational reasoning has improved performance in other research areas. Relational reasoning has been applied to object detection~\cite{hu2018relation} and show that modelling relationships could help object classification and non-maximum suppression. Relational reasoning has also been explored in image captioning~\cite{yao2018exploring} using graph neural networks. Non-local network~\cite{wang2018non} shows that modelling relationship across video frames can significantly boost video classification accuracy.
\subsection{Attention-based Approaches for VQA}
Attention-based approaches have been extensively studied for VQA. Many relational reasoning approaches using attention mechanisms to aggregate contextual information. Soft and hard attention~\cite{xu2015show} has been first proposed by Xu \etal, which has become the main-stream in VQA systems. Yang \etal \cite{yang2016stacked} proposed to stack several layers of attention to gradually focus on the most important regions. Lu \etal~\cite{lu2016hierarchical} proposed co-attention-based methods, which can aggregate information from the other modality. Vaswani \etal ~\cite{vaswani2017attention} aggregated information inside each modality for solving machine translation. Nguyen \etal ~\cite{nguyen2018improved} proposed a densely connected co-attention mechanism for VQA. Bilinear Attention Network~\cite{kim2018bilinear} generated attention weights by capturing the interactions between each feature channel. Structured attention~\cite{zhu2017structured} added a Markov Random Field (MRF) model over the spatial attention map for modelling spatial importance. Besides VQA, Chen \etal ~\cite{chen2017sca} proposed spatial-wise and channel-wise attention mechanisms, which can modulate information flow spatial-wise and channel-wise for image captioning. In referring expression, Xihui \etal~\cite{liu2019improving} propose attention guided feature erasing.
\subsection{Dynamic Parameter Prediction}
Dynamic parameter prediction (DPP) propose another direction for multi-modality feature fusion. Noh \etal ~\cite{noh2016image} firstly proposed a DPP-based multi-modality fusion approach by predicting the weights of fully connected layer using question features. Perez \etal ~\cite{perez2018film} achieved competitive VQA performance compared with complex reasoning approaches on the CLEVR~\cite{johnson2017clevr} dataset by predicting the normalisation parameter of visual features. Furthermore, Gao \etal ~\cite{gao2018question} proposed to modulate visual features by predicting convolution kernels from the input question. Hybrid convolution was proposed to reduce the number of parameters without hindering the overall performance. Beyond VQA, DPP-based approaches have been adopted for transfer learning between classification and segmentation~\cite{hu2018learning}.

\begin{figure*}[t]
        \begin{center}
                \includegraphics[width=\linewidth]{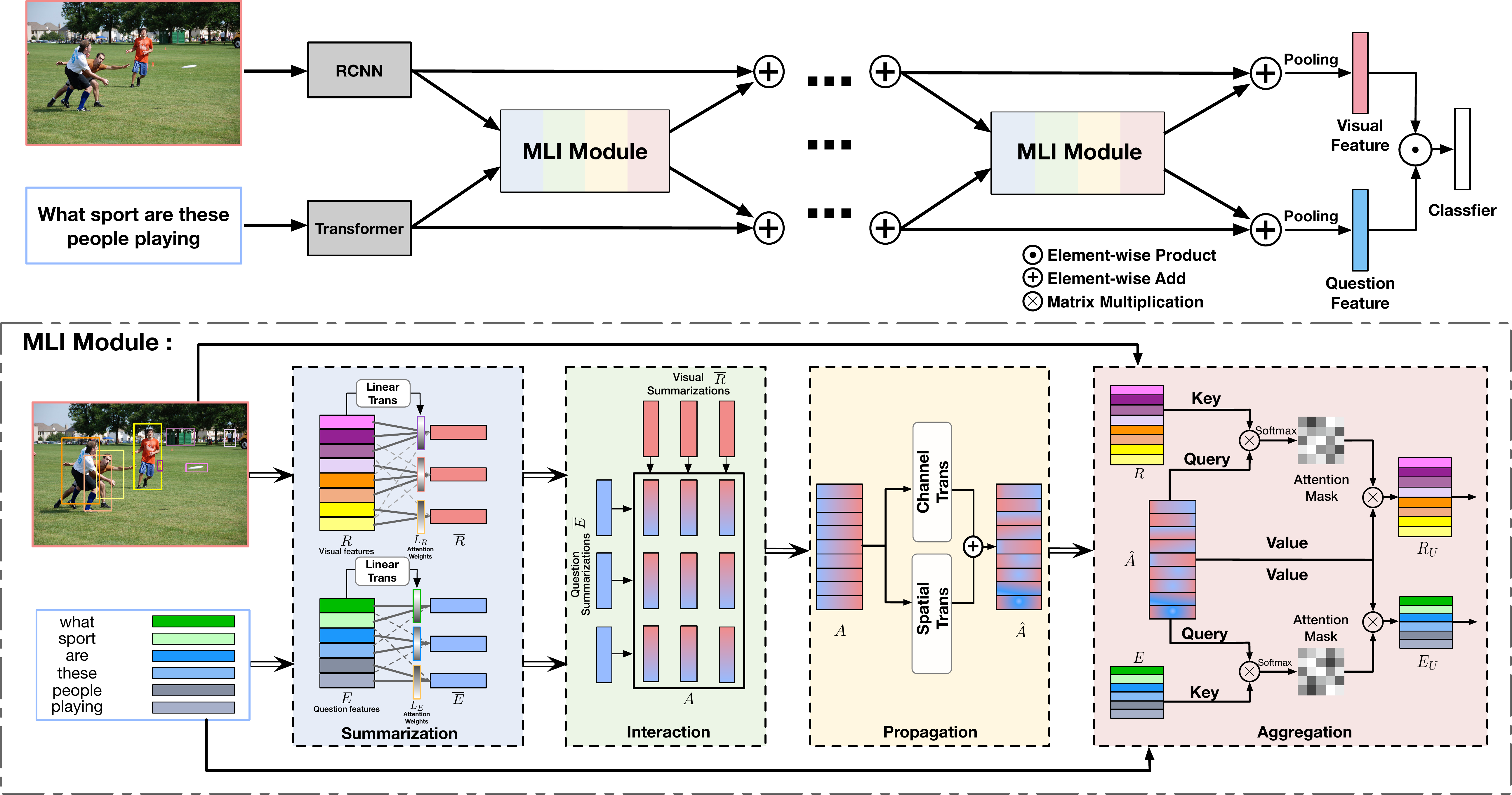}
        \end{center}
        \caption{An overview of our proposed stack Multi-modality Latent Interaction Network. Multi-modality reasoning is accomplished inside our proposed MLI modules. After MLI module, residual connection is used for stacking multiple MLI modules. Inside MLI, visual and question features will be summarised into a few summarization vectors, which are fused to create question and visual summarization pairs. After acquiring latent interaction features, we propagate information between latent summarization pairs. After feature propagation, each question and visual feature will gather information from latent summarization vectors using key-query attention mechanism.}
        \label{fig:overall}
\end{figure*}

\section{Multi-modality Latent Interaction Network}

Figure~\ref{fig:overall} illustrate the overall pipeline of our proposed Multi-modality Latent Interaction Network (MLIN). The proposed MLIN consists of a series of stacking Multi-modality Latent (MLI) modules, which aims to summarize input visual-region and question-word information into a small number of latent summarization vectors for each modality. The key idea is to propagate visual and language information among the latent summarization vectors to model the complex cross-modality interactions from global perspectives. After information propagation among the latent interaction summarization vectors, visual-region and word features would aggregate information from the cross-domain summarizations to update their features. The inputs and outputs of the MLI module has the same dimensions and the overall network stacks the MLI module for multiple stages to gradually refine the visual and language features. In the last stage, we conduct elementwise multiplication between the average features of visual regions and question words to predict the final answer.

\subsection{Question and Visual Feature Encoding}
Given an input image $I$ and a question $Q$, the task of VQA requires joint reasoning over the multi-modal information to estimate an answer. Following previous approaches \cite{anderson2018bottom, kim2018bilinear, peng2018dynamic}, we extract visual-region features from $I$ using the Faster RCNN object detector \cite{ren2015faster, Jiang_Gao_Guo_Zhang_Xiang_Pan_2019} and the word features from $Q$ using a bidirectional Transformer model \cite{vaswani2017attention}. The feature extraction stage is shown in the upper part of Figure~\ref{fig:overall}. Each image will be encoded as a series of $M$ visual-region features, denoted as $R \in \mathbb{R}^{ {M} \times {512} },$ while sentence will be padded to a  maximum length of 14 and be encoded by bidirectional Transformer with random initialization, denoted as $E \in \mathbb{R}^{ {N} \times {512} }$. The multi-modal feature encoding can be formulated as
\begin{align}
     R &= \text{RCNN}(I; \theta_\text{RCNN}), \\
     E &= \text{Transformer} (Q; \theta_\text{Transformer}), \nonumber
\end{align}
where $\theta_\text{RCNN}$ and $\theta_\text{Transformer}$ denote the network parameters for visual and language feature encoding.

\subsection{Modality Summarizations in MLI Module}
Summarization module can be seen from the Summarization part of Figure~\ref{fig:overall}. After acquiring visual and question features, we add a lightweight neural network to generate $k$ sets of latent visual or language summarization vectors for each modality. The $k$ sets of linear combination weights are first generated via
\begin{align}
        L_R = \text{softmax}_{\leftrightarrow}(W_R R^T + b_R),\\
        L_E = \text{softmax}_{\leftrightarrow}(W_E E^T + b_E),
\end{align}
where $W_R, W_E \in \mathbb{R}^{k\times 512}$ and $b_R, b_E\in \mathbb{R}^{k}$ are the $k$ sets of learnable linear transformation weights for each of the modality, and ``softmax$_{\leftrightarrow}$'' denotes the softmax operation along the horizontal dimension.
The individual visual and word features, $R$ and $E$, can then be converted into $k$ latent summarization vectors, $\overline{R}\in \mathbb{R}^{k \times 512}$ and $\overline{E} \in \mathbb{R}^{k \times 512}$, for the visual and language modalities,
\begin{align}
        \overline{R} = L_R \cdot R,\\
        \overline{E} = L_E \cdot E.
\end{align}

Each of the $k$ latent visual or language summarization vectors (\ie, each row of $\overline{R}$ or $\overline{E}$) is a linear combination of the input individual features, which is able to better capture high-level information compared with individual region-level or word-level features. The $k$ summarization vectors in each modality can capture $k$ different aspects of the input features from global perspectives.

\subsection{Relational Learning on Multi-modality Latent Summarizations}
\noindent{\bf Relational Latent Summarizations.}
Relational latent summarization is in correspondence with the Interaction part of Figure~\ref{fig:overall} .The obtained latent summarization vectors encode high-level information from one of the modalties. To reason the correct answer corresponding to the input image and question, it is important to understand the complex cross-domain relations between the inputs. 
We therefore propose to utilize a relation learning network to establish the associations across domains. Motivated by the simple relation network \cite{santoro2017simple}, we create $k \times k$ latent visual-question feature pairs from the above introduced $k$ latent summarization vectors, $\overline{R}$ and $\overline{E}$, in the two modalities. Such $k\times k$ pairs can be represented as a 3D relation tensor $A \in \mathbb{R}^{k\times k \times 512}$:
\begin{align}
        A(i,j,:) = W_A [\overline{R}(i,:) \odot \overline{E}(j,:)] + b_A 
\end{align}
where ``$\odot$'' denotes elementwise multiplication, $W_A \in \mathbb{R}^{512\times 512}, b_A \in\mathbb{R}^{512}$ are the linear transformation parameters that further transforms the cross-domain features.

\noindent \textbf{Relational Modeling and Propagation.}
It is important to propagate information across the two modalities to learn complex relations for answer prediction. Based on our cross-modality relation tensor $A$, we introduce two operations that passes and aggregate information between the paired features. Before information propagation, the tensor $A\in \mathbb{R}^{k\times k\times 512}$ is reshaped to $\tilde{A} \in \mathbb{R}^{k^2 \times 512}$.
 The first cross-modal message passing operation performs an additional linear transformation on each paired feature,
\begin{align}
        \tilde{A}_c = \tilde{A} \cdot W_c + b_c \label{eq1}
\end{align}
where $W_c \in \mathbb{R}^{512\times 512}$ and $b_c \in \mathbb{R}^{512}$ are the relation linear transformation parameters that transforms each paired feature $A(i,j,:)$ into a new 512-dimensional feature.
The second cross-modal information propagation operation performs information passing between different paired features. The $k\times k=36$ paired cross-modal features pass messages to each other, which can be considered as ``second-order'' information for learning even higher non-linear cross-modal relations,
\begin{align}
        \tilde{A}_p = W_p \cdot \tilde{A} + b_p \label{eq2}
\end{align}
where $W_p \in \mathbb{R}^{36\times 36}$ and $b_p \in\mathbb{R}^{36}$ are the linear transformation parameters that propagates information across paired features.
The results of the two cross-modal transformations focus on different aspects of the cross-modal paired features to model the complex relations between the input image and question. The first operation focuses on modeling the relation between each individual visual-question latent pair, while the second operation tries to propagate higher-order information between all visual-question pairs to model more complex relations. 
The summation of the results of the two above operations $\hat{A} \in\mathbb{R}^{k^2\times 512}$,
\begin{align}
        \hat{A} = \tilde{A}_c + \tilde{A}_p \label{eq3}
\end{align}
can be considered as a latent representation that deeply encodes the cross-domain relations between the latent summarization vectors in the two modalities.

\noindent \textbf{Feature Aggregation.} 
The latent multi-modality representation $\hat{A} \in\mathbb{R}^{k^2\times 512}$ contains fused question and region features. Each original visual feature $R(i,:)$ and word feature $E(i,:)$ can aggregate information from the latent representations $\hat{A}$ for improving their feature discriminativeness, which has paramount impact on final VQA accuracy. 
The feature aggregation process can be modeled by the key-query attention mechanism from Transformer \cite{vaswani2017attention}. Each of the region and word features, \ie, $R, E \in \mathbb{R}^{M \text{or} N\times 512}$, would be converted to $128$-d query features, $Q_R, Q_E \in \mathbb{R}^{M \text{or} N \times 128}$, as
\begin{align}
Q_R &=  R \cdot W_{qr} + b_{qr},~~~~E_Q = E \cdot W_{qe} + b_{qe} 
\end{align}
where $W_{qr}, W_{qe} \in \mathbb{R}^{512\times 128}$, $b_{qr}, b_{qe}\in \mathbb{R}^{512\times 128}$ are the linear transformation parameters for calculating the query features. Each feature of the latent representations, \ie, $\hat{A} \in \mathbb{R}^{k^2 \times 512}$, would be converted to $128$-d key and value features $K, V \in \mathbb{R}^{k^2\times 128}$, 
\begin{align}
        K = \hat{A} \cdot W_k + b_k,~~~V = \hat{A} \cdot W_v + b_v,
\end{align}
where $W_k, W_v \in\mathbb{R}^{512 \times 128}, b_k, b_v \in \mathbb{R}^{128}$ are the linear transformation parameters that calculate the key and value features from latent representations $\hat{A}$. 
The query features of the region and word features, $Q_R, Q_E$, would be used to weight different entries from latent representations with their key features $K$,
\begin{align}
        U_R = \text{softmax}_{\updownarrow} \left(\frac{Q_R\cdot K^T}{\sqrt{\text{dim.}}} \right),\\
        U_E = \text{softmax}_{\updownarrow} \left(\frac{Q_E\cdot K^T}{\sqrt{\text{dim.}}} \right),
\end{align}
where softmax$_{\updownarrow}$ denotes conducting softmax operation along the vertical dimension and ``dim.'' $=128$ is a normalization constant. $U_R, U_E\in\mathbb{R}^{M \text{or} N \times k^2}$ stores each region or word feature's weights to aggregate the $k^2$ latent representations. The original region and word features can therefore be updated as
\begin{align}
        R_U = R + U_R \cdot \hat{A}\\
        E_U = E + U_E \cdot \hat{A}
\end{align}
where $U_R \cdot \hat{A}$ and $U_E \cdot \hat{A}$ aggregate the informamtion from the latent representations to obtain the updated region and word features  $R_U$ and $E_U$. The feature aggregation process has been illustrated in the Aggregation module in Figure ~\ref{fig:overall}.

The input features $R, E$ and output features $R_U, E_U$ of the above introduced MLI module shares the same dimension. Motivated by previous approaches \cite{kim2018bilinear, peng2018dynamic}, we stack MLI modules for multiple stages to recursively refine the visual and language features. After several stages of MLI modules, we average pool the visual and word features separately and elementwisely multiplicate the deeply refined region and word features for multi-modal feature fusion. A final linear classifier ($W_{cls}, b_{cls}$ as parameters) with softmax non-linearity function is adopted for answer prediction,
\begin{align}
     R_{\text{pool}} &= \frac{1}{M}\sum_{i=1}^M R_U(i,:), \\
     E_{\text{pool}} &= \frac{1}{N} \sum_{i=1}^N E_U(i,:), \\
     \text{Answer} &= \textnormal{Classifier}~[R_{\text{pool}} \odot E_{\text{pool}}]
\end{align}
Accordingly, the overall system is trained in an end-to-end manner with cross-entropy loss function.
\subsection{Comparison of Message Passing Complexity}
In this section, we compared the message passing complexity between co-attention ~\cite{lu2016hierarchical}, self-attention ~\cite{vaswani2017attention} and intra-inter attention~\cite{peng2018dynamic}. The information flow pattern has been illustrated in Figure~\ref{fig:comparison}. For co-attention, the number of message passings is $\mathcal{O}(2 \times M \times N)$ because each word would calculate an attention matrix from each visual region and vice versa. For self-attention, the number of message passings is $\mathcal{O}(M \times M + N \times N)$. The number of message passings for intra- and inter-modality attention is the summation of those of self-attention and co-attention, $\mathcal{O}((M+N) \times (M+N))$. Generally, in bottom-up \& top-down attention~\cite{anderson2018bottom}, 100 region proposals would be used for multi-modal feature fusion. The quadratic number of message passings in self attention~\cite{vaswani2017attention} and intra- and inter-modality attention flow~\cite{gao2019dynamic} would requires large GPU memories and hinders the relational learning as well. For our proposed MLIN framework, the MLI module generates $k$ latent summarization vectors for each modality. After relational reasoning, $k \times k$ features are generated. In the final feature redistribution stage, $\mathcal{O}(k \times k \times N)$ message passings are performed for question feature update, and $\mathcal{O}(k \times k \times M)$ message passings are required for updating region features. The total number of message passings for our proposed MLIN in each stage is therefore $\mathcal{O}(k \times k \times (M + N))$. Our proposed multi-modality latent representations could better capture multi-modality interactions with much fewer message passings and achieved competitive performance compared with DFAF. A performance comparison has been conducted in the experiments session.

\section{Experiments}

\subsection{Dataset}
We conduct experiments on VQA v2.0~\cite{antol2015vqa} and TDIUC~\cite{kafle2017analysis} datasets. Both VQA v2.0 and TDIUC contain question-image pairs collected from Microsoft COCO~\cite{lin2014microsoft} dataset and annotated questions. VQA v2.0 is an updated version of VQA v1.0 by reducing data bias. VQA v2.0 contains train, validation and test-standards and 25$\%$ of test-standards serve as the test-dev set. Performance evaluation on VQA v2.0 includes evaluating accuracies of different types of questions: YES/NO, Number, Others and overall accuracy. Train, validation and test sets contain 82,743, 40,504 and 81,434 images, with 443,757, 214,354 and 447,793 questions, respectively. We carry out extensive ablation studies on the validation set of VQA v2.0 trained on train split. Also, we report final performance on VQA v2.0 test set trained on the combination of train and validation set, which is a common practice of most previous approaches listed in Table~\ref{tab:vqa}. Although VQA v2.0 has been commonly adopted as the most important benchmark on VQA. However, Kafke \etal ~\cite{kafle2017analysis} found that the performance of VQA v2.0 is dominated by simple questions, which make it difficult to compare different approaches. To solve the bias problem existing in VQA v2.0, TDIUC collect 1.6 million questions divided into 12 categories.

\subsection{Experimental Setup}
We use common feature extraction, preprocessing and loss function as most previous approaches listed in Table~\ref{tab:vqa}. For visual features, we extract the first 100 region proposals with dimension of 2048 for VQA v2.0. While on TDIUC, we extract the first 36 region features. Region features are generated by Faster RCNN~\cite{ren2015faster}. For the question encoder, we pad all questions with 0 to a maximum length of 14 and extract $\mathbb{R}^{14 \times 786}$ question features using a single layer Bidirectional Transformer~\cite{vaswani2017attention} with random initialization. After acquiring visual and word features, we transform them into 512 dimension using linear transform. For all layers, we use a dropout rate 0.1 and clip the gradients to 0.25. Default batch size is 512 with Adamax~\cite{kingma2014adam} optimiser with a learning rate of 0.005. We gradually increase the learning rate to 0,005 in the first 1000 iterations because our Bidirectional Transformer Encoder is initialised randomly while previous approaches use pretrained Glove~\cite{pennington2014glove} and Skipthought~\cite{kiros2015skip} embedding. We also augment our MLIN with a Masked Word Prediction for transformer regularisation. We trained the model for 7 epochs and decay the learning rate 0.0005 and fix it for the following epochs. All layers are initialised randomly with Pytorch's~\cite{paszke2017automatic} random initialisation. For pretrained language models, we adopt a base BERT ~\cite{devlin2018bert} model which is trained by randomly masking words. 

\begin{table}[tb]
\small
\centering
\begin{tabular}{l  c  c}
\toprule
Component & Setting & Accuracy \\
\toprule
Bottom-up~\cite{anderson2018bottom} & Bottom-up & 63.37 \\
\hline
\multirow{3}{2cm}{Bilinear Attention~\cite{kim2018bilinear}} & BAN-1 & 65.36 \\
& BAN-4 & 65.81 \\
& BAN-12 & 66.04 \\
\hline
\multirow{3}{2cm}{DFAF~\cite{peng2018dynamic}} &DFAF-1 & 66.21 \\
&DFAF-8 & 66.66 \\
&DFAF-8 + BERT & 67.23 \\
\hline
\multirow{2}{*}{Default}& MLI-1 & 66.04 \\
       & MLI-8 + BERT & \textbf{67.83} \\
\hline
\multirow{2}{2cm}{\# of stacked blocks} & MLI-5 & 66.32 \\
 & MLI-8 & \underline{\textbf{66.53}} \\
\hline
\multirow{4}{2cm}{\# of Question and Visual Summary Heads} & 3 by 3 &  65.63 \\
 & 6 by 6 & \underline{66.04} \\
 & 6 by 12 & 66.15 \\
 & 12 by 12 & \textbf{66.21} \\
\hline
\multirow{4}{2.5cm}{Latent Interaction Operator} & Concat & 65.99  \\
 & Product &  \underline{66.04} \\
 & Addition &  65.69 \\
 & MUTAN &  \textbf{66.20} \\
\hline
\multirow{2}{2cm}{Embedding dimension} & 512 & \underline{66.04} \\
&  1024 & \textbf{66.18} \\
\hline
\multirow{3}{2cm}{Latent Propagation Operator} & Linear & \textbf{66.04} \\
& Self Attention & 65.84\\
& Dual Attention & 66.01\\
\hline
\multirow{2}{2.5cm}{Feature Gathering Operator} & Key-query & \underline{\textbf{66.04}} \\
& \multirow{1}{*}{Transpose} & \multirow{1}{*}{65.78}\\
\hline
\multirow{1}{3cm}{\# of Parallel Heads in Feature Gathering Operator} & 8 heads & 65.84\\
& 12 heads & \underline{\textbf{66.04}}\\
& 16 heads & 66.19 \\
\hline
\multirow{4}{3cm}{BERT Finetuning} & Freezing & 65.51 \\
& lr 1/10 fintuning & \underline{\textbf{67.83}} \\
& lr 1/100 finetuning & 66.99 \\
& lr 1/1000 finetuning & 66.74 \\
\bottomrule
\end{tabular}
\caption{Ablation studies of our proposed MLIN on VQA v2.0 validation dataset. Default setting is represented by underline while best performance will be highlighted. Our proposed MLIN takes both simplicity and performance into consideration. }
\label{tab:ablation}
\end{table}

\subsection{Ablation Study on VQA2 Validation}
We carried out extensive ablation studies on evaluating the effectiveness of each module in our proposed MLIN in Table ~\ref{tab:ablation}. The default setting is one stage MLIN where all features are transformed into dimension of 512. We create 6 summarizations for each modality. For the feature aggregation key-query attention module, we adopted a 12 head multi-head attention with each head calculating 128-dimensional features. In ablation study, we check the influence of the number of MLIN stacks, number of latent summarisation vectors, latent interaction, latent propagation, feature aggregation and final feature fusion operator. 

Similarly with BAN~\cite{kim2018bilinear} and DFAF~\cite{peng2018dynamic}, we stack the proposed MLI module for 5 and 8 times denoted as MLIN-5 and MLIN-8 for multiple stage reasoning. We observe that deeper layers will improve the performance and can be optimized by SGD thanks to the residual connections~\cite{he2016deep}. 

Then we study the influence of the number of question and visual summarization vectors. Too few summarization vectors will be unable to capture different aspects of the input which deteriorates the overall performance. Too many summarization vectors will require too much GPU memory and computations with marginal improvement. We choose 6 question summarization and 6 visual summarization vectors as a trade-off between performance and computation.

For the interaction operator to create paired summarization vectors, we compare between element-wise product, element-wise addition and bilinear fusion (MUTAN)~\cite{ben2017mutan} for multi-modality summarization fusion. Bilinear fusion~\cite{ben2017mutan} gives the best performance. However, we choose elementwise product in our final model considering the overall simplicity and efficiency of the network design. Different from our approaches, Simple Relational Reasoning Network ~\cite{santoro2017simple} choose concatenation by default.

For the simplicity of hyper-parameter selection, we set all layers have the same dimension. Extracted visual and question features are transformed into the same dimension by linear transform. 1024 leads to better performance than 512. However, stacking multiple MLI modules can lead to more performance improvement than being wide. Our final model chooses 512 dimensions by default. 

Among the latent paired summarization vectors, there exist several ways for propagating information between them. Self-attention~\cite{vaswani2017attention} uses key-query attention to aggregate information from the other latent summarizations. while dual attention aggregate information inside and outside each feature vector simultaneously using self attention.  In our experiment, our proposed relational propagation operations (e.g.\ Equation~\ref{eq1},\ref{eq2},\ref{eq3}) could achieve better performance than the complicated dual attention.

After acquiring latent interaction features, the original question and visual features will gather information from the latent vectors to complete multi-modality relational learning. We tested two approaches for feature gathering from latent vectors. We use the key of visual and word feature to gather information from the query of latent vectors and perform weighted pooling of latent summarization vectors. Motivated by the dynamic attention weight prediction network ~\cite{wu2019pay}, we use the the transpose of attention weight in the summarization stage to gather information from latent summarization vectors. Key-query attention approach outperform dynamic attention weight prediction.

Another hyper-parameter in feature gathering stage is the number of attention heads and head dimension in the feature aggregation stage, we keep the dimension of each heads as 128 and test the number of parallel attention head with number of 8, 12 and 16. The obtained features of different heads are concatenated to obtain the final features.

Language model has been actively investigated in NLP related tasks. Language models~\cite{mikolov2013distributed, pennington2014glove, peters2018deep, devlin2018bert} can generate feature that better capture language meanings. BERT~\cite{devlin2018bert} is a language model pretrained by randomly masking a word or predicting whether one sentence is next to the other sentence. As can be seen from the table, finetuning the MLIN+BERT model by setting its learning rate to 1/10 of the main learning rate will awaken the full power of BERT.

\begin{table}[tb]
\footnotesize
\centering


\begin{tabular}{l c c c c c}
\toprule
\multirow{2}{*}{Model} &  \multicolumn{4}{c}{test-dev} & \multicolumn{1}{c}{test-std} \\
\cmidrule(lr){2-5} \cmidrule(lr){6-6} 
 & Y/N  & No. & Other & All & All\\
\hline
Feature Fusion  & & & & & \\
\hline
BUTP (CVPR)~\cite{anderson2018bottom} & 81.82 & 44.21 & 56.05 & 65.32  & 65.67 \\
MFH (ICCV)~\cite{balanced_vqa_v2}& n/a & n/a & n/a & 66.12 & n/a \\
MFH+BUTD (ICCV)~\cite{balanced_vqa_v2} &  84.27 &  49.56 & 59.89 & 68.76 & n/a \\
BAN+Glove (NIPS)~\cite{kim2018bilinear} & 85.46 & 50.66 & \textbf{60.50} & 69.66 & n/a \\
\hline
Relation Learning & & & & & \\
\hline
DCN (CVPR)~\cite{nguyen2018improved} & 83.51  & 46.61  & 57.26  & 66.87 & 66.97 \\
Relation Prior(Arxiv)~\cite{yang2018multi} & 82.39 &  45.93 & 56.46 & 65.94 &  66.17\\
Graph (NIPS)~\cite{norcliffe2018learning} & 82.91 &  47.13 &   56.22 & n/a &  66.18 \\
Counter (ICLR)~\cite{zhang2018learning} & 83.14 & 51.62 &  58.97 & 68.09 &  68.41 \\
DFAF (CVPR)~\cite{zhang2018learning}&  86.09  & 53.32& 60.49 &70.22 &70.34 \\
DFAF-BERT(CVPR)~\cite{zhang2018learning} &  86.73  & 52.92& \textbf{61.04} &70.59 &70.81 \\
\hline
MLIN(ours) &   85.96  & 52.93 & 60.40 &70.18 &70.28 \\
MLIN-BERT(ours)  &  \textbf{87.07}  & \textbf{53.39}& 60.49 &\textbf{71.09} &\textbf{71.27} \\
\bottomrule
\end{tabular}
\vspace{-6pt}
\caption{Comparison with previous state-of-the-art methods on VQA 2.0 test dataset.}
\label{tab:vqa}
\vspace{-9pt}
\end{table}



   

     


\begin{figure*}[t]
        \begin{center}
                \includegraphics[width=\linewidth]{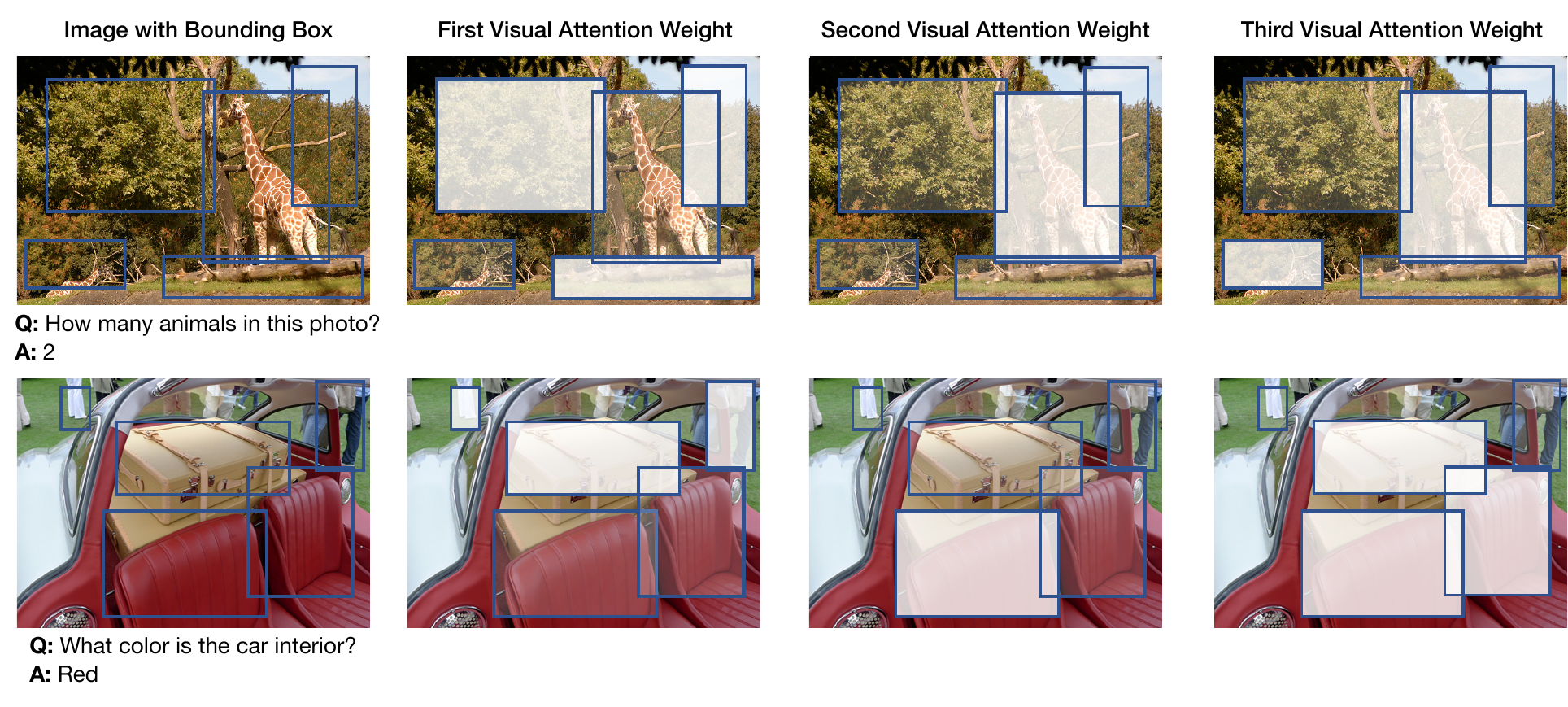}
        \end{center}
        \caption{We visualize the first three visual attention weights for creating visual summarization vectors. Bounding boxes generated by Faster RCNN are shown in the first column. For visual summarization, the colors ranging from clear to white in bounding boxes denote the attention weights from 0 to 1. After training, the first attention focuses on the background regions. The second and the third attention weights concentrate on single and multiple foreground objects}
        \label{fig:vis}
\end{figure*}

\subsection{Comparison with State of the art methods}
In this section, we compare our proposed MLIN with previous state-of-the-art methods on VQA v2.0 and TDIUC datasets in Table ~\ref{tab:vqa} and ~\ref{tab:vqa1}. Following previous methods, we compare our methods on VQA v2.0 test dataset trained with train, validation split and visual genome augmentation.

On VQA v2.0, we divide previous approaches into non-relational and relational approaches which are two orthogonal research directions and can assist each other. Bottom-Up-Top-Down(BUTD)~\cite{anderson2018bottom} approach proposed to use object detection features in a simple attention module for answering the question related to the input image. MFH~\cite{yu2018beyond} is the state-of-the-art bilinear fusion approach. By switching from Residual features to Bottom-up-top-down features, better accuracy can be achieved. BAN~\cite{kim2018bilinear} proposed a bilinear attention mechanism which generates a multi-modality attention using information of each channel and has won the first place in the single model task of VQA competition 2018. 

Besides feature fusion, relational reasoning has been paid much attention in solving VQA. DCN~\cite{nguyen2018improved} proposed a densely connected co-attention module for cross-modality feature learning. $<subject, predicate, object>$ triples are created for VQA reasoning in Relation prior~\cite{yang2018multi}. Conditional Graph~\cite{norcliffe2018learning} built a graph among all region proposals and condition this graph on visual question. Although Conditional Graph is less competitive compared with other approaches. However, the interpretation from conditional graph is quite useful for diagnosing VQA problem. Counter~\cite{zhang2018learning} dives into the number question of VQA by utilising the relative position between bounding box for learning efficient Non Maximum Suppression(NMS). DFAF~\cite{peng2018dynamic} is a multi-layer stacked network by combining intra- and inter- modality information flow for feature fusion. Furthermore, DFAF can dynamically modulate the intra modality information flow using the average pooled features from the other modality. MLI use 100 region proposals for fair comparison.

VQA 2.0 has been mostly adopted as the most important benchmark in VQA. Since VQA 2.0 is dominated by simple samples, which is hard to discriminate between different methods. We also compare with approaches on the TDIUC dataset. QTA~\cite{shi2018question} is the state-of-the-art methods on TDIUC, which proposed a question type guided attention with both bottom-up-top-down features and residual features. Our proposed MLIN can achieve better performance even with bottom-up-top-down features only. Our method also outperform DFAF on this dataset.

\begin{table}[tb]
\footnotesize
\centering

\begin{tabular}{l c c c c | c}
\toprule
\multirow{1}{*}{Model} & RAU~\cite{noh2016training} & MCB~\cite{fukui2016multimodal} & QTA~\cite{shi2018question} & DFAF~\cite{peng2018dynamic} & MLI\\
\hline
Accuracy & 84.26 & 81.86 & 85.03 & 85.55 & 87.60 \\
\bottomrule
\end{tabular}

\vspace{-6pt}
\caption{Comparison with previous state-of-the-art methods on TDIUC test dataset.}
\label{tab:vqa1}
\vspace{-9pt}
\end{table}

\subsection{Visualization}
We visualize the attention weight of summarization vector in Figure ~\ref{fig:vis}. We discover the following patterns. Different summarization have a specific function.
As can be seen from the visualization of attention weight, different summarization vectors focus on different global information. The first attention weight collect information from the background, while the second attention weight focuses on the most important regions for answering the question. While the third attention performs weighted pooling of regions with a strong interaction for answering the question.

\section{Conclusion}
In this paper, we proposed a novel MLIN for exploring relationship for solving VQA. Inside MLIN, multi-modality reasoning is realized through the process of Summarisation, Interaction, Propagation and Aggregation. MLIN can be stacked several layers for better relationship reasoning. Our method achieved competitive performance on benchmark VQA dataset with much smaller message passing times. Furthermore, we show a good pre-trained language model question encoder is important for VQA performance.

\section{Acknowledgements}
This work is supported in part by SenseTime Group Limited, in part by the General Research Fund through the Research Grants Council of Hong Kong under Grants CUHK14202217, CUHK14203118, CUHK14205615, CUHK14207814, CUHK14213616, CUHK14208417, CUHK14239816, in part by CUHK Direct Grant.

{\small
\bibliographystyle{ieee_fullname}
\bibliography{egbib}
}

\end{document}